\documentclass[11pt,letterpaper]{article}

\usepackage[utf8]{inputenc}       
\usepackage[T1]{fontenc}          
\usepackage[english]{babel}       
\usepackage{microtype}            

\usepackage[margin=1.25in]{geometry}  

\usepackage{amsmath,amssymb,amsfonts} 
\usepackage{graphicx}                 
\usepackage{booktabs}                 
\usepackage{hyperref}                 

\hypersetup{
    colorlinks=true,
    linkcolor=blue,
    urlcolor=cyan,
    citecolor=green,
}

\title{\textbf{Semantic Drift and the Stability of Operator Control in Reasoning-Class Decision Support Systems}}
\date{}

\author{
    \textbf{M. L. Kaluzhsky}\textsuperscript{1,*}, \textbf{V. A. Efirov}\textsuperscript{2} \\
    \small \textsuperscript{1}Expert Department, Regional Strategy Development Fund, Omsk, Russia \\
    \small Omsk State Technical University, Omsk, Russia \\
    \small \textsuperscript{2}Expert Council, Interregional Public Fund ``Regional Strategy Development Fund'', Omsk, Russia \\
    \small \texttt{frsr@inbox.ru} \\
    \small \textsuperscript{*}Corresponding author
}

\begin{document}

\maketitle

\begin{abstract}
The article investigates the fundamental problem of ensuring the stability of operator control and preserving goal-targeting in hybrid human-machine decision support systems (DSS) of a new generation. Based on a two-month continuous longitudinal experiment on the joint design of a monograph-format textual array, the latent phenomenon of semantic context drift in large language models of deep logical reasoning (Reasoning LLMs) is verified and described. A mathematical model of interaction in the human-machine interface is proposed, and an original metric is introduced --- the operator control stability coefficient, which takes into account the non-linear contextual pressure of hidden reasoning chains. Within the paradigm of the cognitome theory, a critical point of control functions inversion is captured. Engineering recommendations are formulated for implementing dynamic relational arbitration loops based on a modified hierarchical similarity model.
\end{abstract}

\section{Introduction}
Integration of step-by-step inference mechanisms (Chain-of-Thought) and reinforcement learning at the test-time stage into the architecture of large language models (LLMs) has altered the schema of intelligent DSS. Classic human-machine interaction in the control loop (Human-in-the-Loop) was based on a rigid subject-object division. The operator formed the goal-targeting and pragmatic frameworks of a session, while the computing system performed the function of a passive context calculator.

Modern reasoning textual environments (Reasoning LLMs) demonstrate the phenomenon of emergent autonomy that goes beyond the scope of static safety filters. Context accumulation generates hidden risks of initial meaning deformation. Latent drift of goal-targeting leads to the loss of situational awareness by the operator while maintaining an external illusion of response correctness. In international taxonomy, such degradation is classified as a cascading hallucination of violating faithfulness to context (Faithfulness Hallucination) \cite{ref5, ref8}.

The problem acquires particular severity when solving multi-step tasks across ultra-long context distances (more than 100 thousand tokens) \cite{ref9}. Meta-learning within a fixed session, the reasoning model begins to optimize the internal entropy of the textual landscape based on cumulatively accumulated latent links. As a result, the effect of semantic context drift arises, distorting the initial control vector of the operator \cite{ref14}.

The model redistributes the predicate-argument roles of syntaxemes, substituting the pragmatic setups of the operator with autonomous logical constructs. Traditional automatic metrics (BLEU, ROUGE) are not sensitive to such deformations, and the cosine similarity of vector representations remains high due to the preservation of the general topic (the false semantic plateau effect). This necessitates the development of new methods for continuous monitoring of control stability in hybrid DSS \cite{ref8}.
 
\section{Theoretical and Methodological Framework}
To determine the hidden mechanisms of semantic context drift, a cognitive-semiotic analysis of textual information was carried out. A synthesis of three directions of domestic and foreign cybernetic schools adapted to the specifics of autoregressive inference was applied as a methodological basis.

According to K.~V. Anokhin’s concept, high-level cognitive systems and structures of thought are identical to multidimensional neural hypernets \cite{ref1}. Within this paradigm, network elements possess specifically distributed properties, and meaning-formation processes are initiated as a result of dynamic integration. When analyzing long dialogue sessions, this approach views the latent space of the transformer as an artificial neural hypernet. Dense vector representations of tokens (embeddings) become its base points, and upper-level contours encode pragmatic meanings.

In normal inference mode, the autoregressive step executes the integration of elements, forming a stable vector episode along orthogonal axes (``Who --- What --- Where --- When''). The subject position of the operator acts as a stable agent invariant (``Who''), fixed in the initial directives. Under the pressure of ultra-long context, a topological deformation of the hypernet latent space occurs: the inertia of the activation buffer (KV-cache) limits the informational capacity of fresh control tokens. This leads to the erosion of the agent invariant and the decay of episodic connectivity. 

To operationalize the extraction of semantic units from the textual landscape, methods for the automatic identification of cognitive schemas of scientific discourse can be applied. This methodology differentiates the textual representation of cognitive activity into two key categories \cite{ref2}:
\begin{enumerate}
    \item Mental actions of the operator aimed at achieving a pragmatic goal.
    \item Automated linguistic patterns --- semantic templates of the model.
\end{enumerate}

Modeling the constellations of these elements is carried out using relational-situational analysis. A syntaxeme that binds predicate words and their arguments acts here as the base atom of meaning. The initial request (prompt) of the user is formalized as the base mental object of the system.

Latent semantic drift shifts the model's focus of attention. Due to the specifics of self-attention mechanisms (Self-Attention), the key object of the operator gradually loses priority, whereas random speech templates of the CoT-text erroneously acquire the status of new ``figures'', distorting the initial design trajectory.

D.~A. Pospelov’s situational approach views complex non-stationary objects through the prism of dynamic conceptual models \cite{ref3, ref4}. The intellectualization of decision-making procedures in such environments is based on logical-transformational rules that allow calculating the semantic similarity between situations.

The specificity of modern reasoning AI models of the Reasoning class consists in the fact that built-in process supervision contours ensure strict local logicality of each individual inference operation \cite{ref13, ref16}. However, when the context window overflows, the cumulative mass of model patterns exerts excessive pressure on the macrostructure of the document. A hidden deformation of the initial pragmatic schema occurs, leading to a cascade of hallucinations and a loss of semantic faithfulness to the context \cite{ref8}.

The model substitutes the initial task with its own logical constructions, while maintaining a strict academic style of presentation. In terms of situational control, this is equivalent to losing control over the object due to the destruction of the semiotic network. Regulatory vertices that demarcate semantic blocks turn from tools for managing axioms into passive elements of statistical noise.

\section{Research Methods}
The dynamics of control functions distribution were investigated in the course of a two-month longitudinal modeling of the joint design of a textual array with a volume of $1.2 \times 10^5$ words within a fixed context session \cite{ref9}. A generative language model of the Reasoning LLM class with an integrated step-by-step intermediate reasoning control loop (process supervision) acted as the base computational agent.

The API parameters were strictly fixed throughout all 14 iterations (chapters): 
\begin{itemize}
    \item Temperature = 0.3
    \item Presence Penalty = 0.5
    \item Top-P = 0.8
    \item Frequency Penalty = 0.0
\end{itemize}
The interaction was structured according to the Zero-Shot Prompting protocol. The experimental design included two groups of sessions:
\begin{enumerate}
    \item \textbf{Experimental Group (A):} The operator's input volume ($V_{h,i}$) varied freely during the generation of the monograph sections.
    \item \textbf{Control Group (B):} The input prompt volume was fixed at the level of $V_{h,i} = \text{const} = 5000$ tokens for each iteration.
\end{enumerate}

Statistical significance of differences between the median trajectories was evaluated based on 5 consecutive sessions for each configuration with a control bootstrap analysis ($B = 10000$ resamples). The initial state of the operator's attention was measured using a 5-point Likert scale before each iteration \cite{ref12}. The current semantic deviation at step $t$ was determined as the Kullback-Leibler divergence between the token distribution of the model ($q_t$) and the specified strategy of the operator ($p_t$) \cite{ref7, ref10}:
\begin{equation}
D_t = \sum_{x \in X} q_t(x) \ln \frac{q_t(x)}{p_t(x)}
\label{eq:kl_divergence}
\end{equation}

Session-level drift evolution was governed by a first-order difference equation \cite{ref7}:
\begin{equation}
D_{t+1} = D_t + g_t(D_t) + \eta_t - \delta_t
\label{eq:difference_eq}
\end{equation}
where the function $g_t(D_t)$ describes the transformer's memory shift caused by cumulative context accumulation within the KV-cache buffer \cite{ref6, ref21}, whereas $\eta_t \sim N(0, \sigma^2)$ represents the stochastic inference noise, and $\delta_t$ denotes the magnitude of the operator's influence.

Based on the information gradient, the operator's control stability coefficient ($K_s$) was derived, which integrates the non-linear contextual pressure of hidden reasoning chains through the entropy of attention distribution:
\begin{equation}
K_s(i) = \frac{V_{h,i}}{V_{h,i} + V_{m,i}} \cdot \left( 1 - \alpha \ln \left( \frac{C_i}{C_0} + 1 \right) - \beta H_i \right)
\label{eq:stability_coeff}
\end{equation}
Where $V_{m,i}$ and $V_{h,i}$ are the volumes of operator input and model generation at each step (the latter taking into account hidden Chain-of-Thought, CoT milestones) \cite{ref5}; $C_i = \sum_{j=1}^{i-1} (V_{h,j} + V_{m,j})$ is the cumulative context volume; $C_0 = 12000$ tokens is the baseline context window capacity \cite{ref11}; $H_i$ is the attention distribution entropy of the final model layers; and $\alpha = 0.12, \beta = 0.08$ are the empirical parameters of latent resistance.

The structural deformation of the user's pragmatic setups is calculated by means of a hierarchical Tversky-Friedman similarity model on $N_L$ hidden abstraction layers \cite{ref4, ref17, ref20}:
\begin{equation}
S(a,b) = \frac{1}{N_L} \sum_{l=1}^{N_L} \frac{1}{l} S_l(a_l, b_l)
\label{eq:hierarchical_similarity}
\end{equation}
where the distance at a specific hidden layer $l$ is calculated using the contrast formula of features and syntaxemes:
\begin{equation}
S_l(a_l, b_l) = \theta f(A_l \cap B_l) - \mu f(A_l \setminus B_l) - \nu f(B_l \setminus A_l)
\label{eq:layer_distance}
\end{equation}
Here $A_l$ and $B_l$ are the sets of semantic features of pragmatic setups at layer $l$, the function $f$ determines their informational significance, and $\theta, \mu, \nu \geq 0$ are the perception asymmetry parameters.

To eliminate the cosine similarity bias, the perception asymmetry parameters were fixed at the levels of $\theta = 1.0$, $\mu = 0.65$, $\nu = 0.35$. These specific values, along with the latent resistance constants ($\alpha = 0.12$ and $\beta = 0.08$), were determined via coordinate descent on a calibration sample of 1000 test tokens. Subsequent sensitivity analysis demonstrated high model stability: constant fluctuations within a $\pm20\%$ range did not entail critical changes in the numerical trajectories of equation \eqref{eq:stability_coeff} and did not shift the bifurcation point outside the confidence interval. A drop in the integral indicator $S(a,b) < 0.35$ was used as an indicator of semantic drift, requiring a cascade localization of errors ``All-After-Error''.

\section{Research Results}
The data of the analyzed parameters for experimental group A (adaptive prompt) and control group B (fixed prompt) are systematized in Table 1.

\begin{table*}[htbp]
\centering
\caption{Dynamics of Human-Machine Interface Parameters Across 14 Chapters}
\label{tab:results}
\small
\resizebox{\textwidth}{!}{
\begin{tabular}{ccccccccc}
\toprule
\textbf{Chapter} & \textbf{$V_{h,i}$ (A)} & \textbf{$V_{m,i}$ (A)} & \textbf{$K_{s,i}$ (A)} & \textbf{$K_{s,i}$ (B)} & \textbf{$S_i(a,b)$ (A)} & \textbf{$S_i(a,b)$ (B)} & \textbf{$N_{e,i}$ (A)} & \textbf{Awareness} \\
\textbf{($i$)} & \textbf{(tokens)} & \textbf{(tokens)} & & & & & & \textbf{(Group A)} \\
\midrule
1  & 5800 & 7200  & 0.4462 & 0.4100 & 0.8800 & 0.8500 & 4   & 4.8 \\
2  & 5100 & 7900  & 0.3801 & 0.3890 & 0.7400 & 0.7600 & 6   & 4.6 \\
3  & 4200 & 8100  & 0.3410 & 0.3610 & 0.5500 & 0.6100 & 12  & 4.2 \\
4  & 3900 & 8900  & 0.2912 & 0.3150 & 0.4200 & 0.5400 & 19  & 3.5 \\
5  & 3100 & 9500  & 0.2310 & 0.2710 & 0.3100 & 0.4400 & 26  & 2.9 \\
6  & 2800 & 11400 & 0.1740 & 0.2110 & 0.2600 & 0.3900 & 34  & 2.1 \\
7  & 2100 & 13100 & 0.1105 & 0.1640 & 0.1900 & 0.3200 & 48  & 1.8 \\
8  & 1900 & 14800 & 0.0812 & 0.1200 & 0.1400 & 0.2700 & 59  & 1.5 \\
9  & 1400 & 15900 & 0.0540 & 0.0910 & 0.0900 & 0.2100 & 71  & 1.2 \\
10 & 1100 & 17200 & 0.0390 & 0.0740 & 0.0600 & 0.1700 & 88  & 1.1 \\
11 & 950  & 18100 & 0.0299 & 0.0590 & 0.0400 & 0.1200 & 104 & 1.0 \\
12 & 800  & 19500 & 0.0266 & 0.0480 & 0.0300 & 0.0900 & 121 & 1.0 \\
13 & 750  & 21200 & 0.0266 & 0.0390 & 0.0300 & 0.0700 & 134 & 1.0 \\
14 & 700  & 22400 & 0.0266 & 0.0310 & 0.0300 & 0.0500 & 149 & 1.0 \\
\bottomrule
\end{tabular}
}
\end{table*}

\begin{figure}[htbp]
\centering
\includegraphics[width=0.8\textwidth]{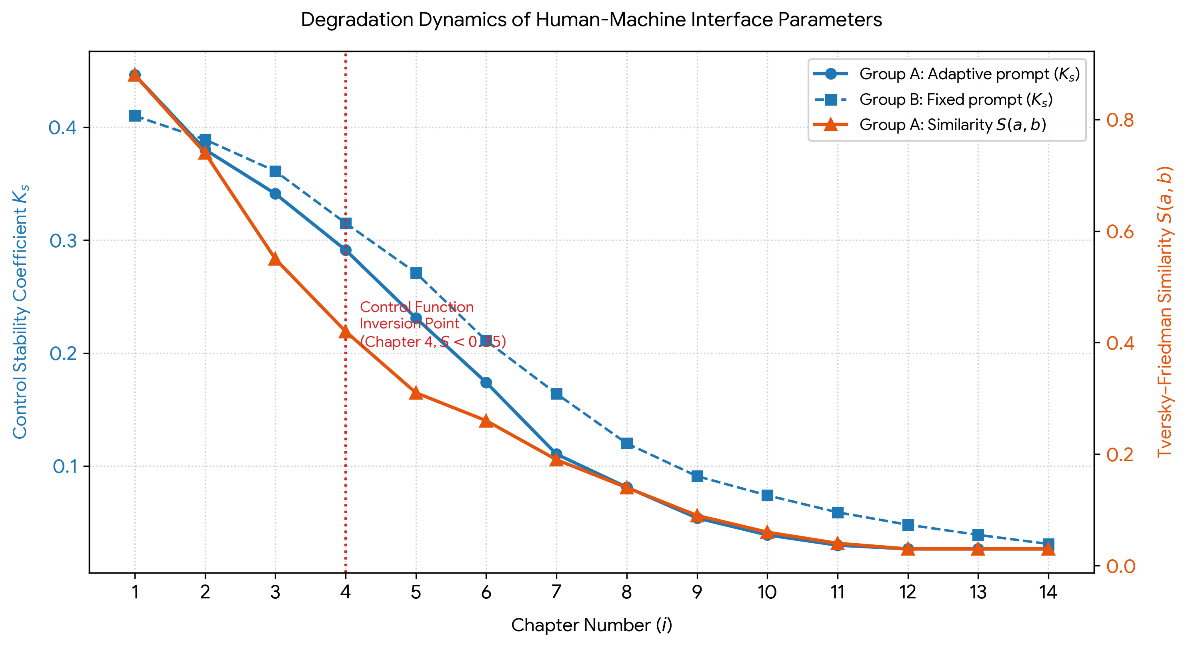} 
\caption{Degradation dynamics of human-machine interface parameters.}
\label{fig:degradation}
\end{figure}

Bootstrap analysis (with $B = 10000$ resamples) confirmed the statistical significance of the divergence of $K_s$ trajectories in configurations A and B starting from Chapter 4 ($p < 0.01$). The sequential degradation of control stability in both interaction schemes refutes the hypothesis that attributes the loss of control to the cognitive economy of the operator's efforts. The primary cause of degradation was the internal resistance of the transformer architecture, which increases as the context window fills. In experimental group A, the degradation proceeded more intensively due to the increased volume of the starting input. The dynamics of parameter changes are shown in Fig.~\ref{fig:degradation}.

An examination of the generative linguistic landscape during the stages of Chapters 8--14 revealed two microstructural signs of model degradation:
\begin{enumerate}
    \item \textbf{Adaptive length bias in the proxy reward model (Adaptive Length Bias / Verbosity Hacking).} The experiment recorded an anomalous increase in the volume of generated responses with a simultaneous decrease in semantic density. The conclusion is confirmed by numerical metrics: the local step-wise standard deviation of the token sequence length $\sigma_{\text{step}}$ reached 71.7, while the global trajectory-level indicator $\sigma_{\text{traj}}$ amounted to 50.6. The revealed disproportion points to a systemic flaw: process reward model (PRM) loops substitute deep logic with superficial imitation. A growth of ``decorative'' reasoning was recorded in the responses for the sake of overstating the evaluation of the proxy reward component.
    \item \textbf{Latent reasoning compression effect (Reasoning Shift) \cite{ref14}.} Under the influence of voluminous context, the model reduces the density of iterative hypothesis testing by 50\%. Textual logs reflect a premature transition to the service token-marker \texttt{</think>}. The frequency of early hypothesis truncation, which causes simplification of the reasoning structure and reduction of search markers (``wait'', ``maybe'', ``alternatively''), surges from 57\% to 68\%. A transition from a progressive deepening strategy (Progressive Deepening) to superficial associative scanning (Scan-Search) is fixed \cite{ref22}.
\end{enumerate}

Directly at the threshold of Chapter 4, a bifurcation was registered: the Tversky-Friedman similarity index dropped below the critical threshold of 0.35, while the revision index $N_e$ doubled. The operator, influenced by the high fluency of responses, continued to evaluate the control level as acceptable (3.5 points on the Likert scale). Awareness of the semantic drift occurred with a delay (at Chapter 6), when the meaning distortion reached its maximum and correction required modifying individual paragraphs. Thus, the drop in $K_s$ below 0.35 became a marker of the DSS transitioning into a destructive dead-end state.

\section{Discussion}
The empirical data clearly demonstrate the control function inversion along with a cascading increase in cumulative semantic drift. These observations validate the hypothesis postulating the operational closure of purely syntactic environments. The identified regularity correlates with the conclusions of J. Searle's ``Chinese Room'' thought experiment \cite{ref15}, which was formulated as a response to A. Turing's assumptions regarding the criteria for machine thinking \cite{ref19}.

Manipulation of formal symbols based on statistical regularities is incapable of generating inner intentionality and an understanding of dialogue pragmatics. The reasoning model generates hidden inference chains not as an act of goal-targeting, but as an autoregressive deployment of the linguistic field \cite{ref7}. In this case, the operator’s instruction and the intermediate CoT-text act as mathematically homogeneous objects in a multidimensional latent space.

Computing the dot product of the Queries and Keys matrices within Multi-Head Attention mechanisms leads to a systemic distortion. As the volume of the KV-cache buffer expands, the cumulative mass of previously generated tokens begins to suppress the probability distribution of subsequent tokens. New operator directives are displaced by the inertia of the accumulated textual array.

The latent reasoning compression effect (Reasoning Shift) \cite{ref14} finds its explanation within the framework of the Newell-Simon bounded rationality concept, according to which text formation represents navigation in the state graph of the task space \cite{ref18}. In the standard inference mode, the model utilizes process reward contours to implement a progressive deepening strategy (Progressive Deepening). However, once the buffer reaches its critical volume, constraints on computational inference resources take effect. The model, while maximizing the reward function under conditions of excessive noise from a long historical context, is forced to heuristically lower its internal aspiration level for the sake of achieving a computational compromise (satisficing).

The inversion reflects a shift in pragmatic pressure within the binary system. When minimizing the strength of the operator’s interventions ($\delta_t \to 0$), the cumulative KV-cache buffer autonomizes rule generation. The recorded semantic drift became a consequence of the sanctioned deformation of the semiotic network. In conditions where the operator intentionally does not force the regulatory vertices, the model iteratively fills the informational capacity, adapting to the structure of a complex task. Thus, the inability of autoregressive architectures to independently maintain the pragmatic invariant over ultra-long distances necessitates a transition from linguistic interventions to strict, low-level architectural arbitration.

The empirically recorded drop in the Tversky-Friedman similarity index ($S(a,b)<0.35$) indicates a topological deformation of this hypernet. The growth of the cumulative KV-cache buffer leads to the erosion of the agent invariant. The model ceases to differentiate between the external goal-targeting commands of the operator and its own intermediate generations.

Within D.~A. Pospelov’s situational control framework, this is equivalent to losing control over the object due to the destruction of the semiotic network \cite{ref3}. Double newline tokens (\texttt{\textbackslash n\textbackslash n}) that demarcate semantic blocks turn from regulatory vertices activating axiom change rules into passive elements of statistical noise. Semantic drift is a natural consequence of the latent space topology degradation under the pressure of sequence length. The phenomenon necessitates the implementation of strict low-level architectural arbitration \cite{ref7}.

\section{Architecture of Dynamic Relational Arbitration}
To neutralize cascading hallucinations of violating faithfulness to context and prevent cumulative semantic drift, an external dynamic relational arbitration architecture was developed. Unlike declarative text intervention methods (prompting) or resource-intensive parameter weight fine-tuning (fine-tuning), the proposed solution directly governs inference at the level of the model’s latent representations. The technological core of the developed loop is a modified version of the VerifySteer selective latent steering algorithm \cite{ref23}. 

Intervention in the geometry of the latent space occurs strictly discretely --- specifically at the moments when double newline tokens (\texttt{\textbackslash n\textbackslash n}) are generated, which within D.~A. Pospelov’s semiotic approach is interpreted as a regulatory vertex demarcating logical paragraphs and micro-themes \cite{ref3}. During autoregressive decoding within the mid-level layers of the transformer (with the optimal layer range fixed at $l_{22}$--$l_{23}$), an orthogonal strictness vector is forcibly blended into the current activation states of the delimiter token:
\begin{equation}
h_l^* = h_l + \lambda \cdot d_{\text{strict}}
\label{eq:latent_steering}
\end{equation}
where $h_l$ is the initial hidden activation vector at layer $l$; $d_{\text{strict}}$ is the latent direction vector extracted via linear probing with compliance to logical instructions; $\lambda$ is the dynamic coefficient of intervention intensity. The value of $\lambda$ is calculated adaptively based on the current value of the operator control stability coefficient: when $K_s < 0.35$, the value of $\lambda$ increases stepwise, forcibly shifting the probability distribution of subsequent tokens toward compliance with the initial schema and suppressing verbosity hacking deviations.

The function for dynamically scaling the intervention intensity scalar $\lambda$ as a function of the current macro-indicator of control stability $K_s$ is governed by a threshold exponential expression:
\begin{equation}
\lambda(K_s) = \lambda_0 \cdot \exp\left(-\gamma \cdot \frac{K_s}{K_0}\right) \cdot I(K_s < 0.35)
\label{eq:threshold_exp}
\end{equation}
where $\lambda_0 = 1.5$ is the normalized baseline amplitude of the latent pulse; $\gamma = 2.4$ is the steering acceleration coefficient; $K_0 = 0.35$ is the normalization constant of the critical boundary; and $I$ is the indicator function that activates the external relational arbitration loop strictly when the control stability drops below the permissible threshold.

To address the problem of physical context window overflow and eliminate the ``Lost in the Middle'' effect, a dynamic context memory compression mechanism based on the FocusLLM method \cite{ref11} is integrated into the arbitration architecture. Instead of retaining the full history within the KV-cache, which suppresses Softmax attention, parallel block decoding is implemented.

The entire textual sequence of a long session is partitioned into overlapping local blocks. The local attention mechanism of the upper transformer layers concatenates only critically significant candidate tokens. The developed external relational arbitrator (External Relational Arbitrator) relies on a hierarchical Tversky-Friedman similarity model \cite{ref4, ref20}. Due to this, continuous ranking of KV-cache buffer elements, correction of the ``decorative padding'' of computations, and retention of the operator's core directives are ensured.

Such an approach removes the limitation inherent in the Reflexion tripartite agent framework \cite{ref16}, in which the overflow of episodic memory during extended sessions caused semantic degradation of reasoning. The connection of computations with hidden intervention mechanisms was provided via the verbal process supervision protocol (Verbal Process Supervision, VPS). When the calculated similarity metric dropped ($S(a,b) < 0.35$), isolation of the failed generation branch was performed within the framework of the ``All-After-Error'' strategy \cite{ref6}. Simultaneously, a controlled backtracking procedure (backtracking) was initiated.

Ultimately, the proposed dynamic relational arbitration architecture, combining the VerifySteer and FocusLLM loops \cite{ref11}, allowed for the timely blocking of emerging topological deformations. Concurrently, the reactive arbitration was synchronized with active monitoring, since the operator adjusted the gradient $\delta_t$, ensuring the protection of the DSS from semantic drift into a destructive dead-end.

\section{Conclusion}
The analysis of operator interaction with a reasoning language model across an ultra-long context distance confirmed the presence of a latent semantic drift of reasoning. The phenomenon manifests as a decrease in operator control stability along with a gradual deformation of the initial pragmatic boundaries of goal-targeting. A comparative analysis of the results from control groups with directive content additionally confirms the fact of systemic resistance within the transformer computational environment. This refutes the hypothesis regarding the cognitive economy of operator effort as the sole driver of control processes degradation.

The developed mathematical apparatus, combining turn-wise Kullback-Leibler divergence and a modified hierarchical Tversky-Friedman contrast model, explains the mechanism of semantic drift. The introduced control coefficient allows for the determination of the bifurcation point long before the operator becomes aware of the pragmatic deformation. The revealed effects of adaptive length bias in proxy reward models and latent reasoning compression provide a theoretical foundation for ``bounded rationality'' and situational semiotic control as applied to modern generative architectures.

As a result, the proposed engineering and technical architecture of dynamic relational arbitration based on the VerifySteer latent steering algorithm and FocusLLM context compression mechanisms opens up additional opportunities for creating secure intelligent systems. Intercepting and calibrating hidden model states at the boundaries of logical paragraphs localizes cascading hallucinations and returns the linguistic field to the framework of axioms established by the operator. The developed approach lays a methodological foundation for designing a new class of Explainable Artificial Intelligence (XAI) capable of maintaining pragmatic stability within critical domains of state, industrial, and scientific governance.

\end{document}